\pdfoutput=1
\documentclass{article}
\usepackage{spconf,amsmath,graphicx}
\usepackage{adjustbox}
\usepackage{cite}
\usepackage{color}
\usepackage{pdfpages}
\usepackage{hyperref}

\title{Interaction-GCN: A Graph Convolutional Network based framework for social interaction recognition in egocentric videos}
\begin{document}
%
\twoauthors
  {Simone Felicioni}
	{University of Perugia\\
	simone.felicioni@studenti.unipg.it
}
 {Mariella Dimiccoli}
	{Institut de Robòtica i Informàtica Industrial (CSIC-UPC)\\
mdimiccoli@iri.upc.edu
\thanks{Work partially funded by projects MINECO/ERDF RyC, PID2019-110977GA-I00, RED2018-102511-T, and MdM-IP-2019-03. }
}

\maketitle
%

%
\begin{abstract}

In this paper we propose a new framework to categorize social interactions in egocentric videos, we named InteractionGCN. 
Our method extracts patterns of relational and non-relational cues at the frame level and uses them to build a relational graph from which the interactional context at the frame level is estimated via a Graph Convolutional Network (GCN) based approach. Then it propagates this context over time, together with first-person motion information, through a Gated Recurrent Unit architecture. Ablation studies and
experimental evaluation on two publicly available datasets validate the proposed approach and establish state of the art results.
\footnote{Features and trained model available at https://github.com/simonefelicioni/InteractionGCN}
\end{abstract}

\begin{keywords}
social interaction recognition, (relational) graph convolutional networks, egocentric vision.
\end{keywords}

%

\section{Introduction}
\label{sec:intro}





\begin{figure}
\centering
\includegraphics[width=0.99\linewidth]{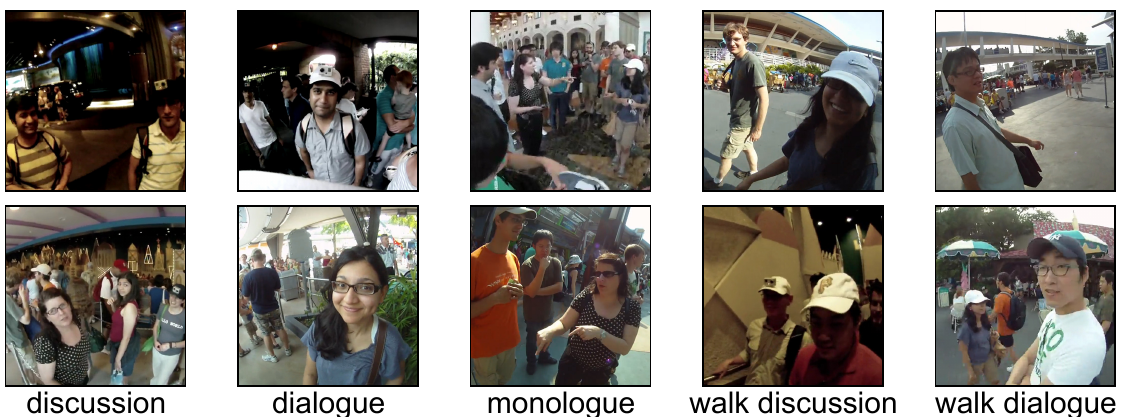}
\vspace{-1mm}
\caption{Different categories in the GeorgiaTech dataset \cite{fathi2012social}}.
\label{fig:classes}
\end{figure}

Recently, wearable cameras have enabled the automatic capture of social life in a naturalistic setting, from a first-person point of view \cite{dimiccoli2018computer}. This has opened the unique opportunity of analyzing the real involvement in social interactions at the personal level \cite{fathi2012social,bano2018multimodal,aghaei2016whom}. 
However, the research focus in the egocentric vision domain has been so far on the detection \cite{bano2018multimodal,aghaei2016whom} and classification of social interactions based on the kind of relations \cite{aimar2019social,aghaei2018towards}, while the problem of fine-grained classification of a specific relation based on the degree of interactivity  has been addressed only in \cite{fathi2012social}. This latter categorization would be of paramount importance to truly  understand social interactions and thus to allow an easier human-machine communication. However, it faces several challenges. Firstly, compared to social relation classification, a categorisation based on the degree of interactivity is more fine-grained and ambiguous. In particular, a model must be able to discriminate whether it is an interactive exchange (discussion or dialogue depending on many people or just two are actively involved, even in presence of multiple persons around) or largely one-sided with a single person speaking most of the time (monologue). 
Secondly, since the camera is worn in a naturalistic setting, interaction cues present a large intra-class variability as well as unpredictable camera/head movement. 
Fig. \ref{fig:classes} illustrates the difficulty of the problem in the GeorgiaTech Social Interaction dataset\footnote{http://cbs.ic.gatech.edu/egocentric/datasets.htm} \cite{fathi2012social}, captured by a head-mounted wearable camera in an amusement park: five different conversation types may occur at different time intervals during a social interaction, even in absence of drastic visual changes, e.g, discussion vs dialogue.
In \cite{fathi2012social}, this problem was addressed by using a Hidden Conditional Random Field (HCRF) formulation that accounts for the dependencies between state labels over time. However, it fails in modelling the interactivity context at the frame level.

\begin{figure}[t!]
\centering
\includegraphics[width=0.99\linewidth]{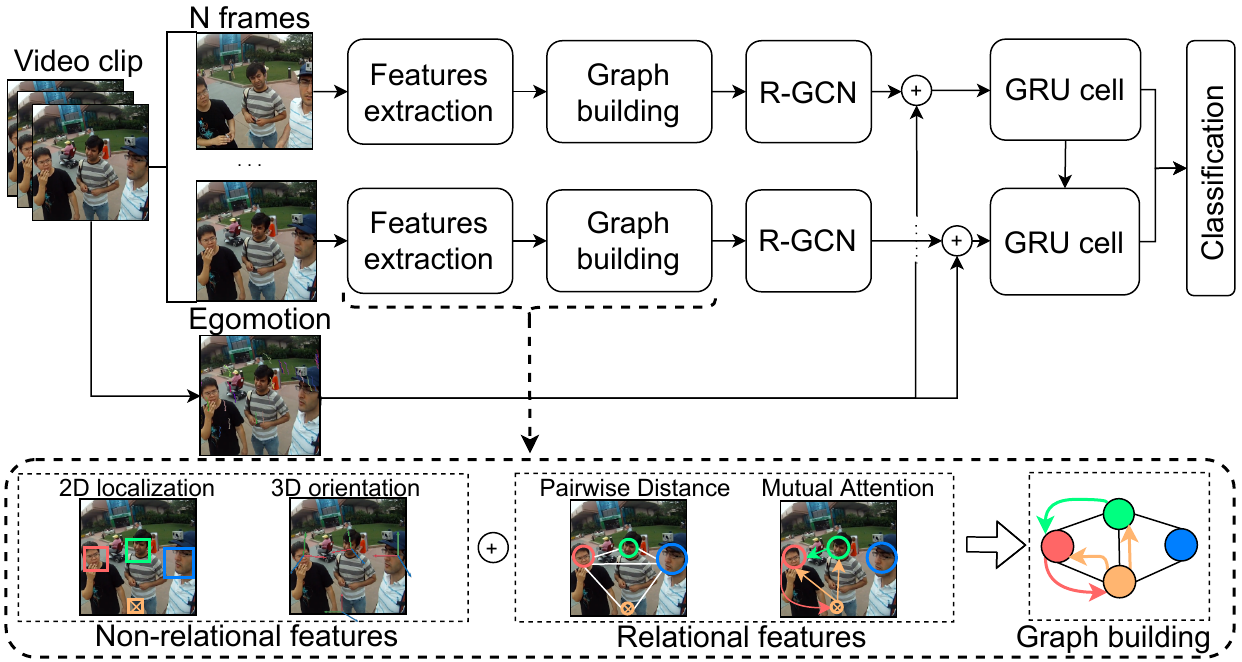} 
   \caption{Overview of the InteractionGCN framework.
   }
\label{fig:workflow}
\end{figure}

In this paper, we overcome this limitation by explicitly modelling the interactivity context by building a relational graph, where nodes correspond to persons with their associated individual features, and edges correspond to interaction cues between pairs of persons. We learn the interactivity context from each frame as embedding on this relational graph via a Relational-Graph convolutional network (R-GCN). It is then fed to a Gated Recurrent Unit (GRU) \cite{cho14learning} together with first-person motion. A visual overview of our InteractionGCN framework is given in Fig. \ref{fig:workflow}.
Experimental results on two publicly available datasets \cite{fathi2012social,aimar2019social} validate the proposed approach. 

\vspace{-6mm}

\section{Related work}
\label{sec:SoA}
\textbf{Third-person social interaction recognition.} A number of works have undertaken the problem of recognizing social relations on still images ~\cite{li2017dual,wang2010seeing},  or particular family relations in personal photo albums ~\cite{dai2015family,xia2012understanding,guo2014graph,chen2012discovering}. More recently, a formalization of people social life based on the Bugental theory has been proposed ~\cite{sun2017domain}.
The literature in the video domain is much more scarce.
A body of work has focused on the detection of \textit{social roles} of people when interacting in videos ~\cite{ramanathan2013social,lan2012social,shu2015joint} by relying on relative age, gender, clothing and again on people relative location. 
Recently,~ \cite{liu2019social} proposed a Multi-scale Spatial-Temporal Reasoning framework to classify social interactions from videos into the eight more common subcategories among the ones proposed in ~\cite{sun2017domain}. 

\textbf{First-person social interaction recognition.} 
So far most efforts have focused on detecting groups of interacting people ~\cite{alletto2015understanding}, detecting social saliency ~\cite{park20123d,soo2015social}, or detecting with whom the camera wearer is interacting ~\cite{aghaei2015towards,aghaei2016whom}.
Only a few works have gone beyond the detection task in the egocentric domain ~\cite{aghaei2018towards,aghaei2017all,aimar2019social}. 
Aghaei  et  al. ~\cite{aghaei2018towards} introduced  a  pipeline  for  automatic  analysis  of  duration,  frequency, type of relation, and diversity of the social interactions of a user captured by a wearable photo-camera during several weeks. A detailed classification of social relations in the egocentric domain, based on the Bugental's theory, has been recently proposed in ~\cite{aimar2019social}.
Similarly to us, ~\cite{fathi2012social} aims at the classification of social interactions into five types  depending on the modality of interaction. 
This is achieved through a HCRF based formulation. Although this model has shown encouraging results, it fails in capturing the conversational context at the frame level.

\textbf{Graph convolutional networks.} 
Recently, GCN have shown promising  results  in  a  variety  of  problems requiring  the  manipulation  of  graph-structured data, including several computer vision tasks \cite{qi20173d,wang2018videos}. The key idea underlying GCN is the propagation of node information through message passing among neighbor nodes that is achieved by aggregating node information. 
Although the success of graph-based modelling for social interaction understanding \cite{fathi2012social,dai2015family,guo2014graph,wang2010seeing,lan2012social,ramanathan2013social}, GCN have been little explored in this context so far \cite{liu2019social}. However, this work focus on recognizing the type of social relation, e.g, colleagues vs father-child for which objects,  and person appearance play an important role. In contrast,  we aim at a more fine-grained classification in terms of dialogue, discussion, monologue, etc. as originally proposed in \cite{fathi2012social}, independently on the kind of relation.

\section{Proposed approach}
\label{sec:approach}
We propose a new approach to classify social interactions in egocentric videos, which follows the taxonomy introduced in \cite{fathi2012social}. The overall architecture of our InteractionGCN framework is shown in Figure \ref{fig:workflow}, and it is detailed below.
\subsection{Feature extraction}
 For each of $M$ persons in the scene, say $[P_1, ..., P_M]$, two different types of features are extracted: non-relational and relational features. The former take into account information that belongs to a specific person, independently of other interacting people. The latter take into account interaction features between pairs of persons. Since the camera is head-mounted, the camera wearer is assumed to look at the center of the scene and to be located at distance zero from the camera. 

\textbf{Non-relational features.} To estimate non-relational features (i.e. 3D head orientation, location, first-person motion), in each selected frame from a video clip we detected faces, grouped them across frames and estimated their 3D orientation and 2D localization by using the Microsoft Azure Cognitive Systems API \footnote{https://azure.microsoft.com/en-us/services/cognitive-services/}. 
Finally, we estimated first-person motion features by computing homography matrices for each pair of consecutive frames as in \cite{jiang2017seeing}.

\textbf{Relational features.}  To estimate directional attention, we assumed that a person's gaze is strongly correlated to his/her 3D head orientation and can be described as a cone in 3D space, so that a person $P_i$ is looking at a person $P_j$  if the cone spanned by the head orientation of $P_i$ includes $P_j$. The relative distance between $P_i$ and $P_j$ was computed by relying on the pre-estimated 2D distance of every person from the camera in a bird-view model. Similarly to \cite{fathi2012social}, the latter was obtained by fitting a polynomial regression model built previously by collecting values of the height of faces in the image at different distance to a tripod mounted GoPro camera.


\subsection{InteractionGCN framework}
We generate a graph $\mathcal{G}=(V, E, R, \mathcal{W})$ for each frame, where $V$ is the set of nodes, $E$ the set of edges, $R$ the set of relations represented by the edges (i.e. distance and attention), $\mathcal{W}$ the set of weights. 
Each node $v_i \in V$ corresponds to a person $P_i$ having as node features non-relational cues $f_i$, e.g. stacked 3D head orientation and 2D localization vectors. Graph nodes are connected by two type of edges $(e_{ij}^r \in E, r \in R= \{d,a\})$, corresponding to two different pairwise relations: relative distance between pairs of individuals in the scene ($d$), including the camera wearer, and directional attention pattern ($a$). Distance edges $(e^d_{ij})$ are bi-directional and represent the distance between $P_i$ and $P_j$, while attention edges $(e^a_{ij})$ are directed edges pointing from $P_i$ to $P_j$, meaning that $P_i$ is looking at $P_j$. $\alpha_{ij}^r \in \mathcal{W}$ is the weight of the labeled edge $e_{ij}^r$.
We feed this graph initialized with the computed relational and non-relational features to a R-GCN \cite{schlichtkrull2018modeling} that propagates information among nodes in a local graph neighborhoods, yielding a richer representation relevant for interaction recognition, as it will be shown in Section \ref{sec:experiments}.
The propagation model used to update a node $v_i$ in a R-GCN is an accumulated  transformed  feature  vectors of neighboring nodes obtained as follows:
$$
h^{(k+1)}_i = \sigma \left( \sum_{r \in \mathcal{R}} \sum_{j \in \mathcal{N}^r_i} \frac{\alpha_{ij}^r}{c_{i,r}} W^{(k)}_r h^{(k)}_j + \alpha_{ii}^rW^{(k)}_0 h^{(k)}_i \right)
$$
where $h^{(k+1)}$ is the hidden  state  of  node $v_i$ at the  $k+1$ layer in a R-GCN, $\sigma$ is the ReLU activation  function, $W_r$ and $W_0$ are learnable parameters of the transformation, $\alpha_{ij}$ and $\alpha_{ii}$ are the edge weights, $\mathcal{N}^r_i$ denotes the set of neighbor indices of node $i$ connected by a relation type $r \in \mathcal{R}$, while $1/c_{i,r}$ is a normalization constant, ensuring that the node $i$  receives a total weight contribution of 1, to which we assigned the value $c_{i,r}= |\mathcal{N}^r_i|$. This model accumulates transformed features of neighboring nodes through a normalized sum. More specifically, the node features are subject to relation-specific transformations, depending on the type and direction of an edge.
Single self-connection of a special relation type to each node in the data  ensures  that the  representation of  a node  at layer $l+ 1$ can also be informed by the corresponding representation at layer $l$. 
The output of the GCN at frame $t$ is a list of node representations $g_t$. This list is vectorized and then concatenated with dynamic features corresponding to first-person motion estimation at frame $t$ in a vector $x_t$. The collection of vectors for each frame is the input of a GRU \cite{cho14learning}, which captures the context information  during all observations. The hidden state $H$ is updated every frame $t$ as follows: $H_{t+1}=GRU(H_t,x_t)$. More precisely:
$$z_t = \sigma(W_{xz}\cdot x_t +W_{Hz} \cdot H_t +b_z) $$
$$r_t = \sigma(W_{xr}\cdot x_t +W_{Hr} \cdot H_t +b_r) $$
$$ \tilde{H} = tanh(W_{xH} \cdot x_t + W_{HH}\cdot (r_t* H_t) +b_H)$$
$$H_{t+1}= z_t * H_t+ (1-z_t) *\tilde{H},$$
where $x_t$ and $H_{t+1}$ are the input and output vectors, $z_t$ and $r_t$ are the update and reset vectors, $W_s$ and $b_s$ are the parameter matrices and bias vectors. The gate vector $z_t$ is the trade-off parameter for updating hidden state $H_{t+1}$ from the previous state $H_t$ and the current estimate $\tilde{H}$.
Finally, the maximum and the average of the output of all timesteps are concatenated and fed to a softmax classifier.

\section{Experiments}
\label{sec:experiments}

\subsection{Experimental setting}
\textbf{Dataset.} In our experiments, we used the GeorgiaTech Social Interaction Dataset \cite{fathi2012social}, consisting of 42 hours of video recorded by 8 different subjects wearing a head-mounted GoPro camera in a Disney World Resort.  
In total there are 1141 annotated video clips, with five types of interaction labels: dialogue, discussion, monologue, walk dialogue and walk discussion. These interactions take place at a dinner table with group of friends, while walking, or while standing in a line, in public transport etc. 
We stress that this is the only available video dataset with conversation type's annotations. Therefore, we further annotated the EgoSocialInteraction dataset \cite{aimar2019social}, including 693 sequences captured by a NarrativeClip worn as a necklace, with the same labels.

\textbf{Evaluation metrics.} Given the unbalanced distribution of labels in the dataset, in addition to the top-1 accuracy, we used also the F-score and we plot the confusion matrix.

\textbf{Implementation details.} In the GeorgiaTech dataset we filtered out individuals with a low level of presence in video shots by computing the percentage of frames in which each person appears in a video shot weighted by the average distance from camera wearer. For the remaining people, when a person moves out of the scene for a few frames, we assume that his/her features could be inferred by interpolation from their corresponding features in temporal adjacent frames. As in \cite{fathi2012social}, we trained our model on five users and tested on three. 
In EgoSocialRelation dataset we found only up to three instance of the \textit{walk} classes and \textit{monologue}. In fact, being the camera worn on the chest instead of the head, people walking side to side are hardly visible in the pictures. Therefore we did not consider these classes in our experiments. 

For both datasets, we performed data augmentation directly on the feature vectors by adding random noise in the direction of the eigenvectors and proportional to the eigenvalues of the feature matrix, multiplied by a Gaussian random variable with zero mean and standard deviation $10^{-4}$ \cite{aghaei2018towards}. We employed a R-GCN with a single layer. Adding layers did not result in further improvements. We performed a grid search over the network’s hyperparameters by training multiple models and choosing the one with best performance on the validation set. The best results were achieved after 83 and 32 epochs for the GeorgiaTech and the EgoSocialRelation datasets respectively, by using a learning rate of $10^{-6}$ , a weight decay of $0.005$, and employing the Adam optimizer with both $L_1$  and $L_2$ regularization.

\textbf{Comparative results.}
In Tab. \ref{tab:comparisons}, we compare the proposed approach to several baseline methods. The first one is \cite{fathi2012social}, that takes as input the same features employed as input to our InteractionGCN model, except the histogram of roles that was computed as in \cite{fathi2012social}, but  relying on the pattern of attention computed as in our approach. We used their HCRF model for classification\footnote{https://github.com/yalesong/hCRF-light} and report the best results obtained with 10 hidden states. In addition, we trained a Multi Layer Perception (MLP) Network, since it has proved to be significantly better than linear models for video classification. 
Furthermore, we implemented the Temporal Graph Convolutional Network (TGCN) model presented in \cite{yan2018spatial}, that connects temporally corresponding nodes but using a RGCN instead of simpler GCN. 
These experiments show that our model achieves a significant improvement over all baseline on both datasets.
 In the EgoSocialRelationDataset we can observe the same performance ranking but with overall better numerical results due the lower number of classes founded in this dataset.
\setlength{\tabcolsep}{1pt}
\begin{table}
\centering
\begin{tabular}{|l|c|c|c|c|c|}
\hline
Model &  \multicolumn{2}{c|}{GeorgiaTech} & \multicolumn{2}{c|}{EgoSocialRelation} \\& Acc & F-score & Acc & F-score \\
 
\hline\hline
HCRF \cite{fathi2012social} & $38.99\%$ & $38.99\%$ & $55.55\%$ & $63.64\%$\\
\hline
MLP & $44.03\%$ & $27.27\%$  & $64.72\%$ & $63.41\%$\\
\hline
T-GCN \cite{yan2018spatial} & $45.67\%$& $44.23\%$ & $81.81\%$ & $81.81\%$\\
\hline
\hline
Ours &  ${\bf 61.88\%}$ & ${\bf 62.36 \%}$ & ${\bf 86.36\%}$ &  ${\bf 84.21\%}$\\
\hline
\end{tabular}
\vskip-3mm 
\caption{Performance comparisons.}
\label{tab:comparisons}
\end{table}

\subsection{Results}
\label{subsec:results}
\begin{table}
\begin{adjustbox}{width=\columnwidth,center}
\setlength{\tabcolsep}{1pt}\vspace{2mm}
\begin{tabular}{|l|c|c|c|c|c|c|c|}
    \hline
    Classes & Discus. &  Dial. & W. Disc. & W. Dial. & Monol. & \multicolumn{2}{|c|}{ All} \\
    \hline\hline
    \#videos & 527 &  421 & 154 & 255 & 74 & \multicolumn{2}{|c|}{ 1431}\\
    \hline\hline
    Removed cue &
    \multicolumn{5}{|c|}{ {\bf Per class accuracy}}&{\bf Accuracy} & {\bf F-score}
     \\
        \hline\hline
    H. localization & $62.5\%$ & $68.6\%$& $58.6\%$ & $ 63.1\%$ & $17.6\%$ &$62.47\%$  & $62.45\%$\\
    H. orientation & $62.5\%$ & $68.6\%$& $55.2\%$ & ${\bf 66.2\%}$ & $23.5\%$ & $ {\bf 62.47\%}$ & ${\bf 62.47\%}$ \\
    M. attention & $61.3\%$ & $60.8\%$& $48.3\%$ & ${\bf 66.2\%}$ & $35.3\%$ &  $59.84\%$ &  $60.34\%$\\
    P. distance & ${\bf63.1\%}$ &  ${\bf70.6\%}$ & ${\bf65.5\%}$ & $55.4\%$ & $23.5\%$ &  $62.20\%$ &  $62.53\%$\\
    F. p. motion & $57.7\%$ & $52.0\%$& $10.3\%$ & $40.0\%$ & ${\bf 41.2\%}$ &   $49.71\%$ &  $48.89\%$\\
    \hline
    \hline
    All cues  &   $57.7\%$ & $ 68.6\%$  &  $62.1\%$ & ${\bf 66.2\%}$ & ${\bf 41.2\%}$ & $ 61.88\%$ & 62.36\% \\ 
    \hline
\end{tabular}
\end{adjustbox}
\caption{\small{Performance by removing an interaction cue at time.}}
\label{tab:ablation2}
\end{table}

\textbf{Ablation study on the architecture.}
To validate the different components of InteractionGCN architecture, we report on Tab.\ref{tab:comparisons} (b) results obtained 1) by using GRU architecture taking as input the extracted features, and 2) by using R-GCN model to get the conversational context at the frame level, that we stacked with first-person motion features to get the class predictions, without GRU. 
These results, obtained on the GeorgiaTech dataset, validate all components of our framework.

\textbf{Ablation study on the features.}
\label{subsec:ablation}
Tab. \ref{tab:ablation2} shows the effect of each feature employed in our model. A large performance drop is experienced when first-person motion is removed. Indeed, first-person motion is specially important to characterize walk classes. However, even without using this feature, we outperform all baselines. Instead, neglecting mutual attention as cue,  specially degrades the classes \textit{monologue} and \textit{walk discussion}.
When removing head localization or head orientation or pairwise distance, the accuracy of the class \textit{monologue} becomes lower than or very close to the chance level, which is totally unsatisfactory for a classification problem with five classes. Instead, by keeping all the features, we overall lose less than $1\%$ globally and up to $5\%$ for the class \textit{discussion}, hence obtaining a more balanced per class accuracy.
 Indeed, during a \textit{monologue} most of people are looking at the person who is talking and at a distance typically larger that the one that characterizes other interactions.


\begin{figure}
\begin{minipage}[c]{0.75\columnwidth}
\centering
\vskip0mm
\hskip-2mm
\includegraphics[width=0.75\columnwidth]{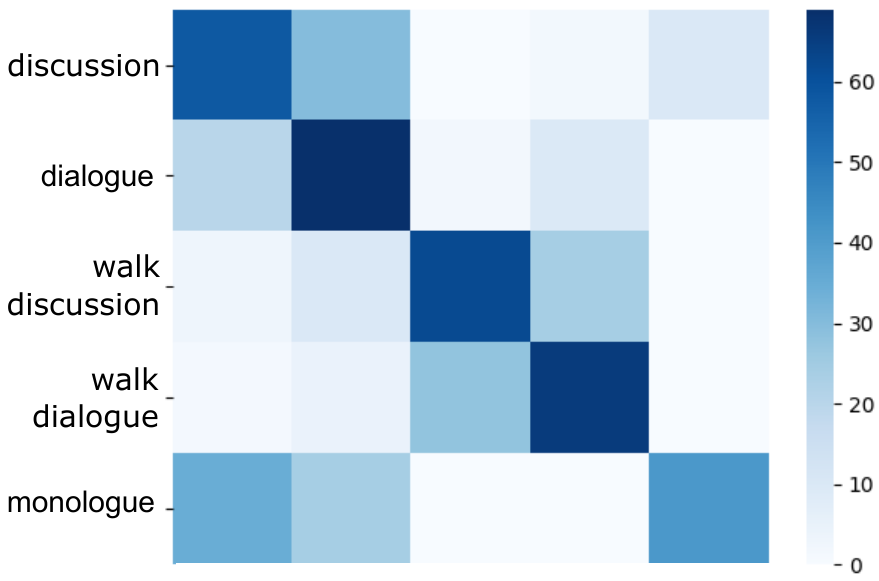}\vskip-2mm
(a)
\end{minipage}\hskip-6mm%
\begin{minipage}[c]{0.25\columnwidth}
\centering\footnotesize
\begin{tabular}{|l|c|c|}
\hline
Model & Acc. &  F-score \\
\hline\hline
GRU & $53.02\%$ & $53.63\%$\\
\hline
R-GCN & $48.56\%$ & $47.56\%$ \\
\hline
\hline
Ours &  ${\bf 61.88\%}$ & ${\bf 62.36\%}$ \\
\hline
\end{tabular}\vskip1mm
{\color{white}.}\hskip6mm(b)
\end{minipage}%
   \caption{\small{(a) Confusion matrix and (b) ablation study.}}
\label{fig:CM}
\end{figure}

%

\textbf{Discussion.} Fig. \ref{fig:CM} shows the confusion matrix obtained with our InteractionGCN framework by using all features. 
Similarly to what was observed in \cite{fathi2012social}, and to a major extent in this work, walk classes are well discriminated from non-walk classes. 
 Contrary to what reported in \cite{fathi2012social}, where (walk) dialogue and (walk) discussion are more significantly confused, in our model the class monologue is the one that is more often misclassified.  
 This is likely due to the fact that our model of attention relies on detected faces in a video clip, whereas the one used in \cite{fathi2012social} does not. 
 Due to the unconstrained nature of the videos, face detection may fail especially when people are wearing accessories such as hats or glasses, hence making this model more prone to erros. Additionally, \textit{monologue} is a class with a considerable smaller number of examples with respect to the others in the dataset. 

\section{Conclusions}
\label{sec:ending}

We presented InteractionGCN, a GCN based framework for categorising social interactions based on the interactivity level in sequences captured by a wearable camera. 
InteractionGCN extracts and models relational and non-relational interaction cues at the frame level through a graph, where people are represented by nodes and pairwise relations between people are expressed through edges. 3D head orientation and 2D localization are employed as node features, while pairwise distance and mutual attention are modeled as edge relations. It first captures the interaction context by a R-GCN on the graph at the frame level, and then feeds this information to a GRU-based architecture, together with first-person motion information.
Experimental results on two public datasets demonstrated the benefits of InteractionGCN over several baselines. 


\clearpage
\bibliographystyle{IEEEbib}
\bibliography{strings,refs}

\end{document}